\documentclass[letterpaper, 10 pt, conference, twoside]{ieeetran}
\usepackage[colorlinks,urlcolor=blue,linkcolor=blue,citecolor=blue]{hyperref}
\usepackage{color,array}
\usepackage{graphicx}
\usepackage{mathptmx}
\usepackage{fancyhdr}
\usepackage{algorithm}
\makeatother
\usepackage{mathtools}
\usepackage[utf8]{inputenc}
\usepackage[table]{xcolor}
\usepackage{tabularx,booktabs}
\newcolumntype{C}{>{\centering\arraybackslash}X} 
\usepackage{cite}
\usepackage{amsmath}
\usepackage{amsmath,amssymb,amsfonts}
\usepackage{graphicx}
\usepackage{textcomp}
\usepackage{xcolor}
\usepackage[flushleft]{threeparttable}
\usepackage{hyperref}
\usepackage{caption}
\usepackage{pifont}
\usepackage{xcolor}
\usepackage{rotating}
\usepackage{rotfloat}
\usepackage{algpseudocode}
\usepackage{gensymb}
\usepackage{textcomp}
\usepackage{placeins}
\usepackage{balance}
\usepackage{subfigure}
\usepackage{booktabs}
\usepackage{tabularx}
\usepackage{array}
\usepackage{pbox}
\usepackage{longtable}
\usepackage{booktabs}
\usepackage[mathscr]{euscript}
\usepackage{amsmath}
\DeclareSymbolFont{rsfs}{U}{rsfs}{m}{n}
\DeclareSymbolFontAlphabet{\mathscrsfs}{rsfs}
\usepackage{xcolor} 

\newtheorem{theorem}{Theorem}
\newtheorem{lemma}[theorem]{Lemma}

\definecolor{MYBLUE}{RGB}{66,164,188}
\definecolor{MYORANGE}{RGB}{240,140,73}
\definecolor{MYGREEN}{RGB}{75,144,102}

\def\BibTeX{{\rm B\kern-.05em{\sc i\kern-.025em b}\kern-.08em
    T\kern-.1667em\lower.7ex\hbox{E}\kern-.125emX}}

\IEEEoverridecommandlockouts                              

\usepackage{graphbox}
\makeatletter
\def\ps@IEEEtitlepagestyle{%
  \def\@oddhead{\mycopyrightnotice}%
  \def\@oddfoot{\hbox{}\@IEEEheaderstyle\leftmark\hfil\thepage}\relax
  \def\@evenhead{\@IEEEheaderstyle\thepage\hfil\leftmark\hbox{}}\relax
  \def\@evenfoot{}%
}
\def\mycopyrightnotice{%
  \begin{minipage}{\textwidth}
  \centering \scriptsize
  This article has been accepted for publication in the IEEE/RSJ International Conference on Intelligent Robots and Systems (IROS) 2025. Copyright~\copyright~20XX IEEE.  Personal use of this material is permitted.  Permission from IEEE must be obtained for all other uses, in any current or future media, including reprinting/republishing this material for advertising or promotional purposes, creating new collective works, for resale or redistribution to servers or lists, or reuse of any copyrighted component of this work in other works.
  \end{minipage}
}
\makeatother

\title{\LARGE \bf
MGPRL: Distributed Multi-Gaussian Processes for Wi-Fi-based Multi-Robot Relative Localization in Large Indoor Environments}

\author{Sai Krishna Ghanta \and Ramviyas Parasuraman 
\thanks{School of Computing, University of Georgia, Athens, GA 30602, USA.}
\thanks{Author emails: {\fontfamily{qcr}\selectfont \{sai.krishna;ramviyas\}@uga.edu}}}

\begin{document}
\maketitle
\newcommand{\revision}{\textcolor{red}}

\begin{abstract}
Relative localization is a crucial capability for multi-robot systems operating in GPS-denied environments. Existing approaches for multi-robot relative localization often depend on costly or short-range sensors like cameras and LiDARs. Consequently, these approaches face challenges such as high computational overhead (e.g., map merging) and difficulties in disjoint environments. To address this limitation, this paper introduces MGPRL, a novel 
distributed framework for multi-robot relative localization using convex-hull of multiple Wi-Fi access points (AP). To accomplish this, we employ co-regionalized multi-output Gaussian Processes for efficient Radio Signal Strength Indicator (RSSI) field prediction and perform uncertainty-aware multi-AP localization, which is further coupled with weighted convex hull-based alignment for robust relative pose estimation. Each robot predicts the RSSI field of the environment by an online scan of APs in its environment, which are utilized for position estimation of multiple APs. To perform relative localization, each robot aligns the convex hull of its predicted AP locations with that of the neighbor robots. This approach is well-suited for devices with limited computational resources and operates solely on widely available Wi-Fi RSSI measurements without necessitating any dedicated pre-calibration or offline fingerprinting. We rigorously evaluate the performance of the proposed MGPRL in ROS simulations and demonstrate it with real-world experiments, comparing it against multiple state-of-the-art approaches. The results showcase that MGPRL outperforms existing methods in terms of localization accuracy and computational efficiency. Finally, we open source MGPRL as a ROS package \url{https://github.com/herolab-uga/MGPRL}.
\end{abstract}


\section{Introduction}
Multi-robot systems (MRS) have gained considerable interest in diverse applications such as search-and-rescue \cite{albanese2021sardo}, logistics \cite{li2021reloc} and environmental monitoring \cite{7347424}. In scenarios where GPS signals are unavailable or when preserving the confidentiality of absolute global positions is critical, robots must rely on relative localization with respect to other robots or environmental markers \cite{latif2023seal}. Effective collaboration, data sharing, and task execution in such environments depend on precise relative positioning. This capability is fundamental for cooperative multi-robot tasks, including rendezvous, formation control, coverage, and path planning, as well as for enabling swarm-level behaviors in robotic systems.

Traditional simultaneous localization and mapping (SLAM) \cite{krishna20233ds} techniques have achieved significant advancements but often rely on computationally intensive sensors like RGB-D cameras and LIDARs, which may not be suitable for resource-constrained robots. Moreover, these solutions can only provide localization when robots navigate across overlapping regions, limiting their applicability in large-scale or disjoint environments. 

To address these challenges, recent research has explored alternative approaches utilizing  Ultra-Wideband (UWB) sensors. Although UWB-based systems \cite{brunacci2023development} provide accurate localization, they require dedicated hardware, extensive calibration, fixed deployment and maintenance, increasing the cost and complexity of implementation \cite{zhou2009rfid}. In contrast, Wi-Fi-based relative localization (RL) offers a cost-effective and scalable alternative, leveraging existing wireless infrastructure. Unlike UWB, Wi-Fi-based techniques utilize received signal strength indicator (RSSI) or channel state information (CSI) \cite{yang2013rssi,xue2019locate,jirkuu2016wifi} to estimate the distance between source and mobile device. These approaches reduce dependency on external infrastructure while enabling localization in environments where GPS or visual SLAM may fail.

\begin{figure}[t]
\centering
 \includegraphics[width=0.99\linewidth]{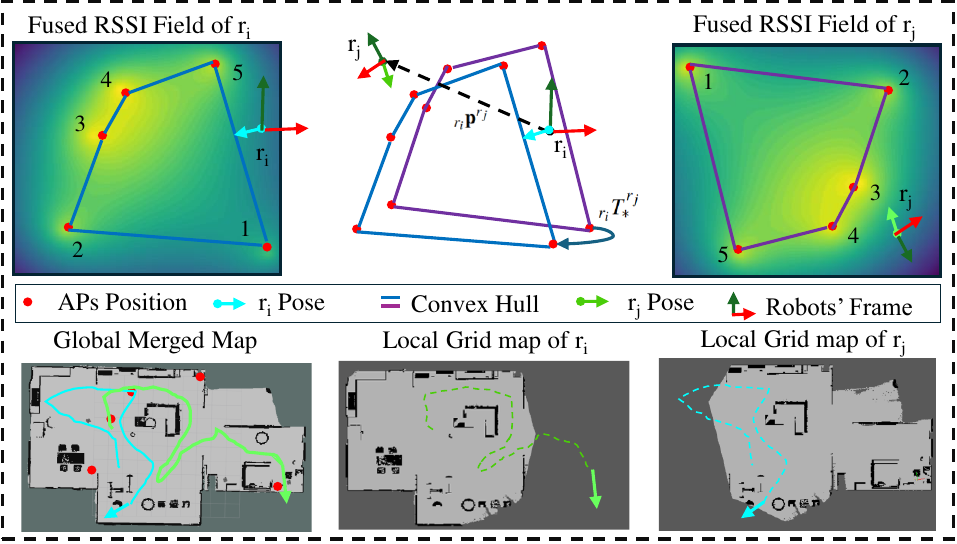}
 \caption{\footnotesize{Illustration of the proposed online MGPRL framework. Each robot identifies multiple common APs via a computationally efficient MOGP pipeline, then leverages convex hull alignment of these APs to achieve relative localization among robots. We predict individual RSSI fields for each AP; the figure displays a fused RSSI field for illustration.}}
 \label{fig:intro}
\end{figure}

The CSI measurements \cite{chen2017confi} are highly sensitive to environmental changes and are not accessible on most Wi-Fi devices \cite{xiao2013pilot}. In contrast, RSSI is widely supported across commercial Wi-Fi hardware, including smartphones and IoT devices, making it more accessible for real-world applications \cite{haseeb2018wisture}. However, RSSI-based localization is affected by signal multipath fading, environmental features, and noise \cite{pandey2022empirical,parasuraman2018kalman}. This necessitates the development of efficient algorithms to model RSSI information for accurate relative localization. 

Though there are RSSI fingerprinting methods \cite{mok2007location}, they require extensive offline data collection and calibration and lack the adaptability and scalability in time-varying and unknown environments. Some works exploited RSSI modeling using spatial modeling and Gaussian Processes Regression (GPR)-based methods to detect the position of Access Points (AP) with respect to the robots \cite{fink2010online,parasuraman2013spatial,xue2019locate}. Moreover, the approaches for robot relative localization rely on unrealistic assumptions. For instance, in a work, HGPRL \cite{latif2024hgp} performs RSSI-based relative localization in MRS, but with a known orientation, which is a major compromise. 

Therefore, we propose a distributed architecture for RSSI-based relative localization that overcomes the limitations of existing works in AP localization and relative localization. The proposed approach introduces a coupled framework based on co-regionalized multi-output GP (MOGP) for simultaneously handling multiple AP localization and integrating it with a novel convex-hull (of APs) based relative localization, as illustrated in Fig. \ref{fig:intro}.  The core novelties and contributions of this paper are:
\begin{itemize}
    \item We develop a computationally efficient co-regionalized MOGP-based RSSI field prediction and AP localization approach that simultaneously models RSSI measurements across multiple APs and performs precise, uncertainty-aware AP localization for each robot. 
    \item We introduce a novel multi-robot relative localization framework that estimates the transformations between the robots based on the geometric alignment of the weighted convex hull of localized APs. The novelty of this approach lies in its scalability and its ability to achieve accurate relative localization, even in resource-constrained robot systems.
\end{itemize}

We evaluate the performance of MGPRL through extensive simulation and real-world experiments and compare its accuracy against the leading multi-robot relative localization technique (HPGRL \cite{latif2024hgp}).
Additionally, we made MGPRL publicly available as a ROS package\footnote{\url{https://github.com/herolab-uga/MGPRL}}, enabling the robotics community to adopt and build upon our framework. A supplementary video showcasing the method in both simulated environments and real-world robotic deployments is attached.

\section{Related Work}
Relative localization in multi-robot systems remains a critical area of research, particularly in GPS-denied environments where accurate positioning is essential for cooperative tasks. In such environments, Wi-Fi signals based learning approaches are being proposed to perform robot localization. For instance, Chen et al. \cite{hsieh2019deep} employed neural networks for indoor localization by leveraging both RSSI and CSI information to provide a flexible system capable of adapting to different environmental conditions. Moreover, Shushuai et al.\cite{li2022self} proposed a self-supervised approach that utilizes monocular vision and lightweight neural networks to estimate the relative positions of multiple robots. Despite their advantages, these neural network-based methods generally depend on large labeled datasets, which are frequently collected through time-consuming fingerprinting procedures involving numerous access points. Moreover, real-time acquisition at every point in environment of all the signal measurements is challenging. Additionally, These approaches tend to overfit to a single environment, limiting generalization—a key concern for neural networks trained on restricted datasets. Considering these challenges, a recent survey \cite{chen2022survey} underscored the requirement of machine learning in multi-robot relative localization and also highlighted the need for more robust and generalizable algorithms.

In contrast, several works have used GPR within active learning frameworks to model Wi-Fi signals for localization, addressing the shortcomings found in offline learning-based techniques\cite{fink2010online}. For example, Xu et al. \cite{xu2014gp} proposed an efficient GP-based approach that relies on sampling only a small fraction of the RSSI data to reduce the computational load in GP-based robot positioning. Meanwhile, Li et al. \cite{hsieh2019deep} proposed a GP-based strategy to calibrate multi-channel RSSI measurements. Moreover, in work \cite{quattrini2020multi}, the authors used GPR to model communication maps (GPRCM) for indoor multi-robot environments. Xue et al. \cite{xue2019locate} introduced a modified error GP regression (MEGPR) method to improve device localization - also adaptable for locating multiple APs - although it requires an extensive offline fingerprinting stage. These GP-based approaches are computationally demanding, especially when dealing with large coverage areas or high-resolution tasks. Furthermore, these methods train separate GP models for each AP, which significantly increases computational complexity and are often infeasible for large-scale environments where the demand for scalability and real-time processing cannot be compromised.

Currently, HGPRL \cite{latif2024hgp} is the most relevant state-of-the-art in online AP-based multi-robot relative localization, using a hierarchical inference-based GP approach to efficiently and accurately estimate an AP position (global maxima) and use it in an algebraic relative localization. HGPRL had outperformed the other existing methods, such as MEGPR \cite{xue2019locate} and GPRCM \cite{quattrini2020multi}, and demonstrated the potential of hierarchical single-output GP predictions over sparse GP and gradient search techniques. However, to remedy the single AP consideration, the authors assumed that the initial robots' orientations are set/known in a global frame to guide relative localization, which is a major limitation in scenarios where robots must be deployed or added in an ad-hoc manner.

\begin{figure*}[t]
    \centering
    \captionsetup{font=small}
    \includegraphics[width = \linewidth]{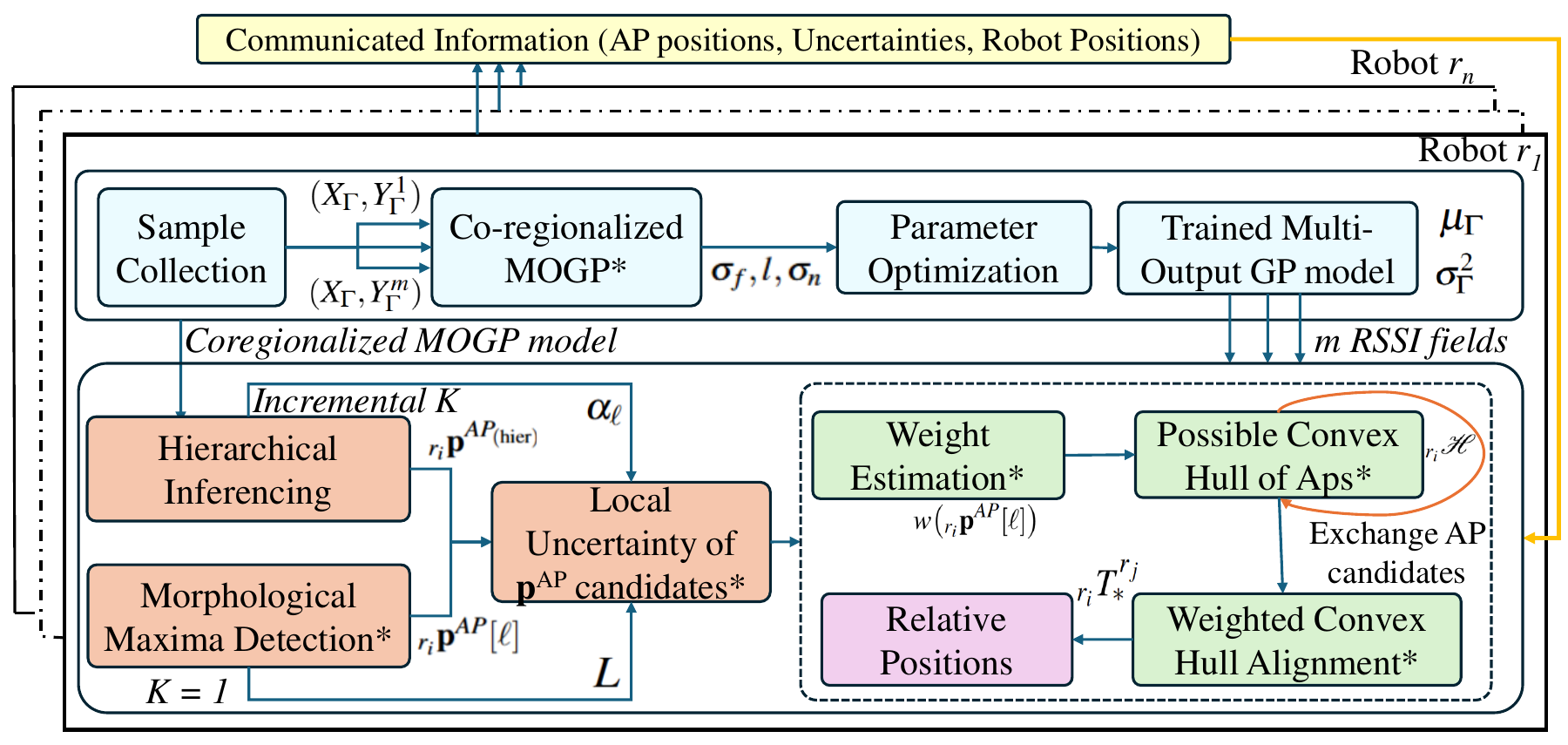}
    \caption{{\footnotesize{Overview of the MGPRL methodology. The \textcolor{MYBLUE}{-light blue} represents \textbf{co-regionalized MOGP} processes for RSSI field prediction, \textcolor{MYORANGE}{-orange} denotes \textbf{uncertainty-aware AP positions estimation}, and \textcolor{MYGREEN}{-green} indicates \textbf{weighted convex hull-based relative localization}. Asterisk (*) elements are newly introduced, integrated approaches—this paper’s novel contributions.}}}
    \label{fig:methodology}
\end{figure*} 
 
All these existing works \cite{latif2024hgp,xue2019locate,quattrini2020multi} treat AP localization and relative localization as separate processes. However, coupling these processes enhances performance by allowing each to compensate for other's errors.
Also, works utilizing GP for RSSI field estimation rely solely on the mean for AP localization without utilizing the variance of GP estimates. However, quantifying the variance can provide essential information to understand the uncertainty of AP position estimation. These works also fail to handle multiple peaks in RSSI distribution, which occur frequently due to environmental conditions such as multipath fading and interference, leading to inaccurate localization in real-time.

We aim to address these limitations by introducing an active learning-based relative localization framework. Our approach, leverages co-regionalized multi-output GP (MOGP), which significantly reduces the computational burden of employing seperate GP model for locating multiple sources. MGPRL overcomes the limitation of multiple maxima by detecting and considering them as candidate AP positions (local maxima), which are further coupled with relative localization mechanism. We model the uncertainty of AP localization and utilize them to couple with the weighted convex-hull-based relative localization mechanism. We accurately transform the positions of neighboring robots into robot reference frames using convex-hull alignment. Moreover, MGPRL offers a distributed and computationally efficient solution, making it well-suited for real-time deployment.

\section{Proposed Methodology}
The proposed MGPRL is a distributed framework, which consists of three key on-board (within robot) processes as shown in Fig. \ref{fig:methodology}: 1.) Co-regionalized MOGP for RSSI environment modeling, 2.) Uncertainty-aware AP positions estimation, and 3.) Weighted convex hull-based relative localization. Each AP's RSSI field is predicted using co-regionalized MOGP, providing mean and uncertainty estimates at every point in the environment. From these RSSI predictions, hierarchical inferencing \cite{latif2024hgp} selects a global maxima, which we refer to as the hierarchical AP position, and morphological maxima detection identify additional candidate positions (local maxima). We then weigh these candidates based on local uncertainty and align their corresponding convex hulls with those received from neighboring robots to estimate inter-robot relative poses.

\subsection{Problem Formulation}
Assume a set of $n$ robots $R\in\{r_1, r_2..., r_n\}$ connected to $m \ge 3$ fixed Wi-Fi APs, denoted as $\mathcal{A} \in \{AP_1, AP_2..., AP_m\}$. Each robot \( r_i \) is equipped with a Wi-Fi receiver and is capable of measuring the RSSI from all APs. Let ${}_{F}p^{r_i}$ represent the position of robot $r_i$ expressed in frame of reference $F$. The measurements collected by the robot $r_i$ at all the sampling points is represented as $\Gamma^{AP}_i$. We perform environment modeling of all APs $\Gamma^{AP_j}_i \in \Gamma^{AP}_i$ separately, to extract mean  $(\mu_{\Gamma_{i}})^{AP_j}$ and uncertainty (variance) $(\sigma^2_{\Gamma_{i}})^{AP_j}$ of each point in environment. These $(\mu_{\Gamma_{i}})^{AP_j}$ and $(\sigma^2_{\Gamma_{i}})^{AP_j}$ are utilized to estimate hierarchical ${}_{r_i}\mathbf{p}^{AP_{\text{(hier)}_j}}$  position  and candidate ${}_{r_i}\mathbf{p}^{AP}_L$ positions. We assume robots exchange their positions, estimated AP ${}_{r}\mathbf{p}^{AP_\text{(hier)}}$, AP candidates ${}_{r}\mathbf{p}^{AP}$ in their \textbf{own frame of reference}. The \textit{primary objective} of proposed approach is to determine the relative positions ${}_{r_i}p^{j}$ of all neighboring robots $R_j \in R \setminus \{r_i\}$ with respect to robot $r_i$.

\subsection{Co-regionalized MOGP for RSSI Field Prediction}
Wi-Fi RSSI measurements have proven effective for radio mapping and localization, as their noisy nature can be modeled using a Gaussian distribution, enabling accurate geospatial\footnote{By jointly modeling multiple APs, MOGP mitigates localized disruptions. For instance, if one AP provides excessively noisy data, correlated signals from the other APs help the model adapt, resulting in more reliable predictions and enhanced uncertainty estimates.} estimation \cite{10649822}\cite{fink2010online} and user positioning \cite{latif2024hgp}. This paper utilizes multi-output Gaussian Processes  \cite{liu2018remarks}, which is able to capture geo-spatial correlations among multiple access points. The MOGP offers improved correlation modeling, sampling complexity, and prediction uncertainty when compared with a single GP used in existing works \cite{xu2019survey}. A MOGP extends traditional GPs to handle multiple correlated output functions and enables joint modeling of these observations.

Let \( X_\Gamma = \{x_1, \dots, x_\gamma\} \) represent the \( \gamma \) 2D planar positions within the environment where a robot has collected RSSI measurements \( Y_\Gamma^{i} = \{y^i_{x_1}, \dots, y^i_{x_\gamma}\} \) from $AP_{i}$, identifiable by its unique MAC address. 
 We define multi-output GP as a collection of correlated latent functions modeled jointly as:
\vspace{-1mm}
\begin{equation}
    \mathbf{f}(x) \sim GP(\mathbf{m}(x), K_s(x, x') \otimes B)
\end{equation}

where \( \mathbf{m}(x) \) is the mean function, and \( K_s(x, x') \otimes B \) is the covariance function $K$ capturing geo-spatial correlations with  the Kronecker product of spatial kernel  \( K_s(x, x') \) (covariance between locations $(x, x')$) and \( B \) (co-regionalization matrix). To model spatial kernel  \( K_s(x, x') \), we utilize Squared Exponential (SE) kernel to prioritize the correlation of proximal locations over distant locations:
\vspace{-1mm}
\begin{equation}
   (K_s)_{x, x^{'}} =  k(x, x') = \sigma_f^2 \exp \left( -\frac{\|x - x'\|^2}{2l^2} \right)
\end{equation}

where \( \sigma_f^2 \) is the signal variance, \( l \) is the length-scale parameter, and \( \|x - x'\|^2 \) represents the squared Euclidean distance between \( x \) and \( x' \). The hyperparameters \({\sigma_f, l}\) are learned  training data by optimizing the log-likelihood of \( Y \) observed RSSI values and \( X \) the corresponding locations:
\vspace{-1mm}
\begin{equation}
\{\sigma_f^*, l^*\} = \texttt{argmax}_{\{\sigma_f, l\}} \texttt{log}  P(Y | X, \{\sigma_f, l\})
\end{equation}

The co-regionalization matrix \( B = [b_{ij}] \in \mathbb{R}^{m \times m} \), positive semi-definite matrix, captures the correlation between multiple APs, where \( b_{ij} \) represents the correlation between $AP_i$ and $AP_j$. After training, the posterior mean and variance of the RSSI prediction for any test location \( q_* \) are given by:
\vspace{-1mm}
\begin{equation}
    \mu_\Gamma[q_*] = m(q) + k_*^T (K + \sigma_n^2 I)^{-1} (y_q - m(q))
\end{equation}
\vspace{-2mm}
\begin{equation}
    \sigma^2_\Gamma[q_*] = k_{**} - k_*^T (K + \sigma_n^2 I)^{-1} k_*
\end{equation}

where, \( K = K_s(X_\Gamma, X_\Gamma) \otimes B\) is covariance matrix between training measurements \( X_\Gamma \), \( k_* = K_s(x_{{q}^*}, X_\Gamma ) \otimes B \) is covariance vector between the test location $x_{{q}^*}$ and training points \( X_\Gamma \), \( k_{**} = K_s(x_{{q}^*}, x_{{q}^*}) \otimes B \) is covariance scalar at the test location $x_{{q}^*}$, \( \sigma_n^2 \) is the noise variance, \( I \) is the identity matrix, $m(q)$ is prior mean function at previous test location $q$. The training complexity of single GPR for all the $m$ APs has  $\mathcal{O}(m\gamma^3)$ and inference complexity $\mathcal{O}(m\gamma^2)$. In contrast, MOGP jointly models each AP measurements with complexity $\mathcal{O}(\gamma^3 + m^2\gamma^2)$ for training and $\mathcal{O}(\gamma^2 + m^2\gamma)$ for inference, improving both performance and efficiency. Further reductions are achievable using sparse MOGP \cite{yang2018online}.

\subsection{Uncertainty-aware AP positions estimation}
\begin{figure}[t]
\centering
 \includegraphics[width=\linewidth]{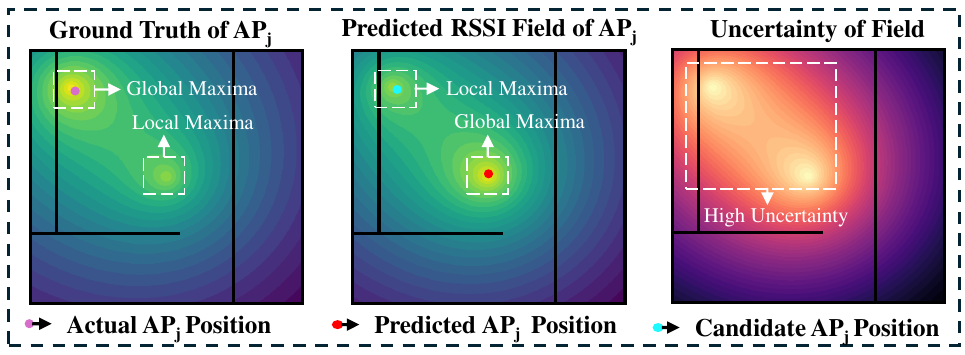}
 \caption{\footnotesize{Motivating example where hierarchical inferencing incorrectly estimate AP position, causing significant relative localization error. High uncertainty is observed in these regions due to small sample size. Our maxima detection selected local maxima (candidate positions).}}
 \label{fig:m1}
\end{figure}

Existing AP localization approaches rely on detecting global maxima in the RSSI distribution. However, RSSI distributions often exhibit multiple peaks due to environmental factors such as constructive interference or sensor noise. This leads to inaccurate AP position estimation, as shown in Fig. \ref{fig:m1}, which in turn affects the precision of relative localization to a great extent. To address these challenges, we leverage existing hierarchical inference and multi-peak detection methods, and introduce an uncertainty-based weighting mechanism alongside a weighted convex hull framework to achieve robust relative localization.

\subsubsection{The AP Position Selection}
We utilize the hierarchical inferencing from \cite{latif2024hgp} to estimate the location of an AP. Let the environment be discretized into cells $\mathbf{c}$. Existing work \cite{latif2024hgp} demonstrates that hierarchical inferencing identifies the global maximum more accurately than other methods. We define a bounded region $\mathbf{B}_v$ at the $v$-th hierarchical level ($v=1,\dots,V$) with progressively finer resolution. For each cell $\mathbf{c}\in \mathbf{B}_v$, we estimate predicted mean RSSI $\mu_{\mathbf{B}_v}(\mathbf{c})$, predictive standard deviation $\sigma^2_{\mathbf{B}_v}(\mathbf{c})$ from trained MOGP. At each hierarchy level, we select the cell with the maximum RSSI (mean) value as the AP location and also as region $\mathbf{B}_{k+1}$ for the next level of resolution to identify the accurate AP position. Eventually, at the last level $V$, we obtain a final AP location ${}_{r_i}\mathbf{p}^{AP_{(\text{hier})}}$. Instead of solely relying on ${}_{r_i}\mathbf{p}^{AP_{(\text{hier})}}$, we capture the maxima, close (w.r.t RSSI strength) to  $AP_\text{(hier)}$, in RSSI field at coarsest level $B_1$ based on morphological maxima detection \cite{serra1994morphological}.  This approach efficiently (quickly) detects local maxima but often fails to identify the global maxima, so we use it exclusively for local maxima identification. We define a local neighborhood $\Omega(\mathbf{c'})$ around each cell $\mathbf{c'}$ and identify number of local maxima appear:
\begin{equation}
L \;=\; \sum_{\mathbf{c'} \in \mathbf{B}_1} 
1\!\Bigl(\,\mu_{\mathbf{B}_1}(\mathbf{c'}) \;=\; 
\max_{\mathbf{u}\,\in\,\Omega(\mathbf{c'})}\,\mu_{\mathbf{B}_1}(\mathbf{u})\Bigr)
\label{eqn:um}
\end{equation}
, where $1(\cdot)$ is the function that counts a cell $\mathbf{c'}$ if it is a local maximum relative to its neighborhood $\Omega(\mathbf{c'})$ and $L$ is number of \emph{detected maxima} in $B_1$. All these maxima are represented as \textit{candidate} AP positions $ {}_{r_i}\mathbf{p}^{AP}_L= \bigl\{{}_{r_i}\mathbf{p}^{AP}[1], \dots,{}_{r_i}\mathbf{p}^{AP}[L]\bigr\}$ for robust AP localization. 



\subsubsection{Weighting Function for AP Positions}
After estimation of all possible candidates ${}_{r_i}\mathbf{p}^{AP}_L$ and ${}_{r_i}\mathbf{p}^{AP_{(\text{hier})}}$, we weigh the candidates based on the local uncertainty. The local uncertainty $U(\mathbf{c''})$ of a cell $\mathbf{c''}$ is defined as the average standard deviation  $\sigma^2_{\mathbf{B}_v}(\mathbf{c''})$, derived from the multi-output GP, over the neighborhood $\Omega(\mathbf{c''})$  (region) around $\mathbf{x}$ at the coarsest level $k=1$ as shown:
\vspace{-1mm}
\begin{equation}
    U(\mathbf{c''})
    \;=\;
    \frac{L}{\lvert \Omega(\mathbf{c''}) \rvert}
    \sum_{\mathbf{u}\,\in\,\Omega(\mathbf{c''})}
    \sqrt{\sigma^2_{\mathbf{B}_1}(\mathbf{c''})})
    \quad 
    \label{eq:Uc}
\end{equation}

For ${}_{r_i}\mathbf{p}^{AP}_L$ and ${}_{r_i}\mathbf{p}^{AP_{(\text{hier})}}$, we assign weights based on local uncertainty of each position:
\vspace{-2mm}
\begin{equation}
    w\bigl({}_{r_i}\mathbf{p}^{AP}[\ell]\bigr)
    \;=\;
    \max\!\Bigl\{\epsilon,\; \frac{1}{1 + U\bigl({}_{r_i}\mathbf{p}^{AP}[\ell]\bigr)}
    \Bigr\},
    \label{eq:weight1}
\end{equation}
\begin{equation}
    w\bigl({}_{r_i}\mathbf{p}^{AP_{(\text{hier})}})
    \;=\;
    \max\!\Bigl\{\epsilon,\;
        \alpha \times \frac{1}{1 + U\bigl({}_{r_i}\mathbf{p}^{AP_{(\text{hier})}})}
    \Bigr\},
    \label{eq:weight}
\end{equation}

where $\ell$ is position in list, $\epsilon>0$ is a small constant to avoid zero weights, $\alpha$ is priority constant, to \textbf{prioritize} the location found by hierarchical inference. The positions with very high uncertainty contribute minimal weight.

\subsection{Weighted Convex Hull-based Relative Localization}
Each robot $r_i$ identifies all locations of APs ${}_{r_i}\mathcal{A}$  from both hierarchical AP localization and morphological maxima detection,   in its own reference frame, forming a multiple convex hull of access points as
\vspace{-1mm}
\begin{equation}
  {}_{r_i}\mathcal{H} \;=\;
  \texttt{ConvexHull}\Bigl(\{{}_{r_i}\mathbf{p}^{AP}[\ell]\}_{\ell=1}^{L+1}\Bigr)
\end{equation}

${}_{r_i}\mathcal{H}$ is the minimal convex polyhedron enclosing all the AP locations in the frame of reference of \( r_i \) respectively. The \{${}_{r_i}\mathbf{p}^{AP}_L$, ${}_{r_i}\mathbf{p}^{AP_{(\text{hier})}}\}$\} (comprising $L+1$ positions) replace the node of $AP_j$ when constructing and optimizing the convex hull.

For two robots $r_i$ and $r_j$, we seek the rigid transformation ${}_{r_i}T^{r_j}\in SE(3)$ that best aligns $r_j$'s AP position sets (${}_{r_i}\mathbf{p}^{AP}_\theta \in $ \{${}_{r_i}\mathbf{p}^{AP}_L$ , ${}_{r_i}\mathbf{p}^{AP_{(\text{hier})}}\}$) to $r_i$'s AP position sets (${}_{r_j}\mathbf{p}^{AP}_\theta \in \{{}_{r_j}\mathbf{p}^{AP}_L$ , ${}_{r_i}\mathbf{p}^{AP_{(\text{hier})}}\}$). Unlike classical trilateration \cite{yang2021survey} or single‐hull alignment, we handle multiple maxima and uncertainties via a \emph{weighted} SVD-based convex hull alignment\footnote{Each robot may create multiple convex hulls for different neighbors (i.e, the convex hull robot $r_i$ for neighbor $r_j$ could be different from robot $r_i$ for neighbor $r_k$ (see Lemma~\ref{lemma:detect})).}:
\begin{equation}
\begin{split}
    {}_{r_i}T_*^{r_j} 
\;=\;
\underset{{}_{r_i}T^{r_j} \in SE(3)}{\arg\min}
\;\;\sum_{AP=1}^{M}\;
\sum_{\ell=1}^{L+1}
w\bigl({}_{r_i}\mathbf{p}^{AP}_\theta[\ell]\bigr)
\,\bigl\|
{}_{r_i}\mathbf{p}^{AP}_\theta[\ell] \\
-
{}_{r_i}T^{r_j} \cdot {}_{r_j}\mathbf{p}^{AP}_\theta[\ell]
\bigr\|^2
\end{split}
\label{eq:weighted}
\end{equation}
, where ${}_{r_i}\mathbf{p}^{AP_j}_\theta[\ell]$ is $\ell$-th AP position for $AP_j$ in hull $r_i$, ${}_{r_j}\mathbf{p}^{AP_j}_\theta[\ell]$ is the corresponding $\ell$-th candidate in frame $r_j$ and $w$ is weight. We utilize Singular Value Decomposition (SVD) based point set alignment \cite{chang2023svdnet} for optimization. The transformation ${}_{r_i}T_*^{r_j}$ is considered true only when the alignment error is less than user-defined $\Lambda$:
\begin{equation}
\label{eq:threshold}
\sum_{AP=1}^M \sum_{\ell=1}^{L+1}
w\bigl({}_{r_i}\mathbf{p}^{AP}_\theta[\ell]\bigr)\,
\bigl\|
{}_{r_i}\mathbf{p}^{AP}_\theta[\ell]
-
{}_{r_i}T_*^{r_j} \cdot {}_{r_j}\mathbf{p}^{AP}_\theta[\ell]
\bigr\|^2
\;<\;
\Lambda
\end{equation}

Once the optimal transformation ${}_{r_i}T_*^{r_j}$ is found, we estimate the relative location of $r_j$ in $r_i$'s frame:
\begin{equation}
    {}_{r_i}\mathbf{p}^{r_j} 
    \;=\;
    {}_{r_i}T_*^{r_j}\cdot {}_{r_j}\mathbf{p}^{r_j},
    \quad
    \text{where} \quad
    {}_{r_j}\mathbf{p}^{r_j}=[\,0,0,0\,]^\top
\end{equation}

If no convex hull alignment is found (i.e., no feasible match exists), the hull is updated with next AP estimates based on the transformation yielding the maximum alignment. The overall time complexity for convex hull-based transformation estimation is $\mathcal{O}(M_{\mathrm{total}} \,\log\,M_{\mathrm{total}})$, where $M_{\mathrm{total}}$ is number of  AP estimates (${}_{r_i}\mathbf{p}^{AP}_\theta + {}_{r_j}\mathbf{p}^{AP}_\theta$). In contrast, the time complexity of traditional trilateration is $\mathcal{O}(M^3)$.


\begin{lemma} \label{lemma:valid}
The transformation estimation remains valid for all types of convex hulls, including worst-case scenarios such as symmetric hulls and collinear APs.
\end{lemma}
\textbf{Proof} We consider three cases to support this lemma. \\
\textbf{Case 1: General Case or Non-Collinear APs:}
When the APs in \({}_{r_i}\mathcal{H}\) and \({}_{r_j}\mathcal{H}\) are non-collinear, the weighted convex hulls formed by robots $r_i$ and $r_j$ encloses distinct spatial structures. As long as there are at least three APs ($m \ge 3$), this approach guarantees stable transformations (\({}_{r_i}T_*^{r_j}\) and \({}_{r_j}T_*^{r_i}\), which have minimum alignment error), unaffected by geometric degeneracies.

\textbf{Case 2: Symmetric Hull:}  
Suppose \({}_{r_i}\mathcal{H}\) and \({}_{r_j}\mathcal{H}\) exhibit rotational symmetry. In the absence of identifiable APs, such symmetry might allow multiple plausible transformations (e.g., reflection across the hull’s symmetry plane). However, as we have unique AP correspondence (i.e., node $AP_i$ of \({}_{r_i}\mathcal{H}\) matches node $AP_i$ of \({}_{r_j}\mathcal{H}\) by MAC address, there exist valid alignment solution with the proposed approach, where one that maps each anchor ${}_{r_i}\mathbf{p}^{AP_i}$ to its \emph{unique} counterpart. This \emph{breaks} any potential rotational ambiguity. Hence, the convex-hull-based alignment returns correct \( {}_{r_i}T_*^{r_j}\).

\textbf{Case 3: Collinear APs:}  
If all APs in \({}_{r_i}\mathcal{H}\) and \({}_{r_j}\mathcal{H}\) lie along (or near) a single line, the shape of each convex hull is degenerate. Nevertheless, as long as each AP is uniquely identified in both frames, this solution will minimize alignment error with respect to \emph{the same} AP correspondences. Geometric degeneracies affect the rank of the covariance matrix in alignment procedure but not the \emph{correctness} of transformation as long as each APs \({}_{r_i}\mathbf{p}^{AP_i}\) pairs only with its \emph{unique} counterpart \({}_{r_j}\mathbf{p}^{AP_i}\) in neighbor robot's hull.

\begin{lemma} \label{lemma:detect}
    The transformations are valid even when a robot \(r_i\) cannot detect certain APs, as long as there are at least 3 unique overlapping APs between two robots.
\end{lemma}

\textbf{Proof}
If \(r_i\) cannot detect few \(AP_u \in \mathcal{A}_{r_j}\), it simply omits \(AP_u\) from its convex hull. The alignment in  \eqref{eq:weighted} only depends on \(|\mathcal{A}_{r_i} \cap \mathcal{A}_{r_j}| \ge 3\), where each AP has a unique MAC correspondence. Hence, \(r_i\) aligns partial hull of shared APs with that of \(r_j\), yielding close transformation \({}_{r_i}T_*^{r_j}\).

\begin{figure*}[t]
\centering
 \includegraphics[width=\linewidth]{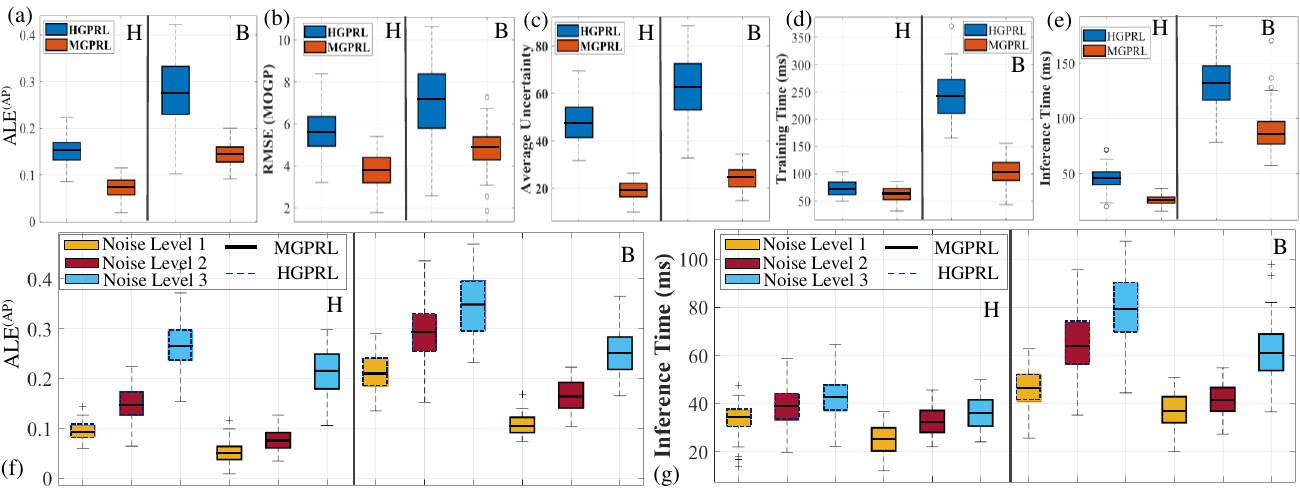}
 \caption{\footnotesize{The performance of AP localization in AWS \textbf{H}ouse ($n=3$) and \textbf{B}ookstore ($n=6$) (marked as H and B): (a), (f) shows the ALE  of AP localization with different noise levels and subfigures (b)-(g) represent RMSE, average uncertainty, inference and training time of the co-regionalized MOGP for RSSI field prediction. Moreover, the inference time of MOGP is analyzed with varying noise levels in (g). }}
 \label{Exp_Box}
\end{figure*}

\section{Experimental Results and Analysis}
We deployed MGPRL into the ROS-\textit{Gazebo} AWS House\footnote{\url{https://github.com/aws-robotics/aws-robomaker-small-house-world}} 
(70$m^{2}$ area) and AWS Bookstore\footnote{\url{https://github.com/aws-robotics/aws-robomaker-bookstore-world}}
 (100$m^{2}$ area) environments using Turtlebot-3 Waffle robots. The MGPRL framework leverages the ROS-SPACE \footnote{\url{https://github.com/herolab-uga/SPACE-MAP}} package for exploring environments. It also consists of frontier detection, navigation and mapping pipelines to explore environment and generate 2D grid maps. We primarily performed two sets of experiments in AWS house and bookstore worlds with $n=3, m = 4$ and $n=3, m = 6$, respectively. We simulated multiple APs by placing them at fixed locations within the simulation environment \cite{ahn2010simulation}. We simulate the RSSI distribution for $AP_j$ based on a path-loss model at position $x$:
\begin{equation}
  \text{RSSI}_j(\mathbf{x}) 
  = 
  R_{d_{0}}
  - 10\,\zeta\,\log_{10}\!\Bigl(\tfrac{d(\mathbf{x},\mathbf{x}_{AP_j})}{d_0}\Bigr)
  +   F_j^{\text{large}}(\mathbf{x}) + F_j^{\text{small}}(\mathbf{x}) ,
\end{equation}

where $d(\mathbf{x},\,\mathbf{x}_{\text{AP}_j})$ is the Euclidean distance, $\zeta=3$ is the path-loss exponent ($2 \le \zeta \le 4$), $R_{d_{0}}=-20dBm$ is the reference power at distance $d_0=1m$, $ F_j^{\text{large}}(\mathbf{x})\sim \mathcal{G}(0, \sigma^2=6dB)$ is a zero-mean, correlated Gaussian random field function $\mathcal{G}(\cdot)$ for log-normal shadowing, and $F_j^{\text{small}}(\mathbf{x}) \sim \mathcal{N}(0, \sigma^2_{small} = 1dB)$ is an random variable function $\mathcal{N}(\cdot)$ for Rayleigh fading.  For measurements $y^j_{x}$ from AP $j$ at location $\mathbf{x}$, we include measurement noise (based on noise level $\Delta$) from random field function $\epsilon_{m,i} \sim \mathcal{N}(0,\sigma^2= \Delta = 0, 1, 2 dB)$ to simulated data:
\begin{equation}
       y_{j,x} \;=\; \text{RSSI}_j(\mathbf{x}) \;+\; \epsilon_{m,i},
   \quad \epsilon_{m,i} \,\sim\, \mathcal{N}(0,\sigma^2) 
\end{equation}
 Initially, the robots collect 15 RSSI samples from each AP while performing a random walk. These initial samples are used to train a co-regionalized MOGP. After each iteration, the GP is updated with new samples, potentially collected at different locations, thereby enhancing its prediction performance. The algorithmic parameters are $\Lambda=0.05m^2$ for strict alignment, which can be increased for relaxed alignment, and $\alpha=1.5$, which can be increased to prioritize hierarchical inference. The performance of MGPRL is compared with HGPRL \cite{latif2024hgp}, which addresses the same problem as ours.
 For a fair comparison, we reimplemented HGPRL for multiple APs (applying a hierarchical single-output GP for every AP) and assumed known robot orientations in the global frame to aid relative localization. This comparison not only provides a head-to-head evaluation of AP localization and relative localization, but also highlights the benefits of our MOGP approach over running separate single-output GPs.

 \begin{figure*}[t]
\centering
 \includegraphics[width=\linewidth]{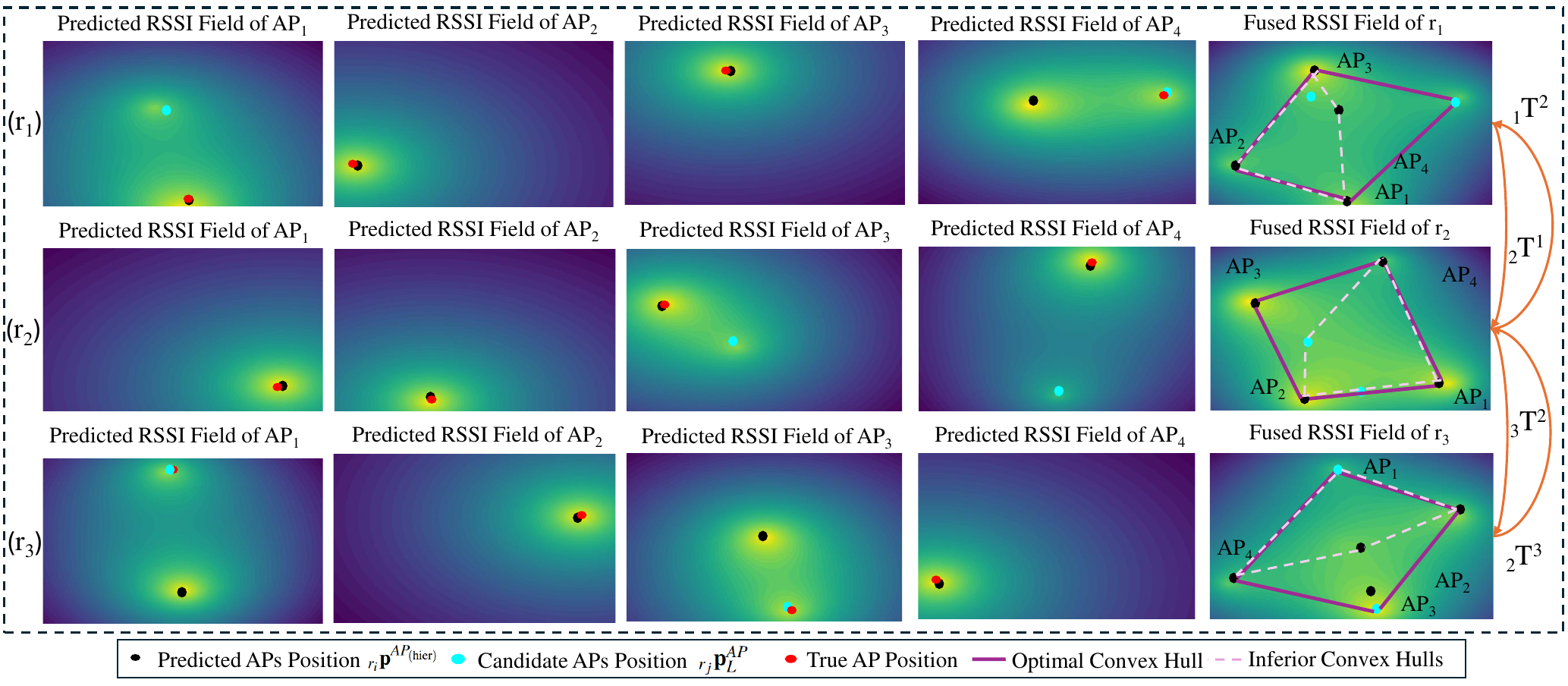}
 \caption{\footnotesize{The predicted RSSI fields for each AP and fused RSSI field in AWS house experiment; The predicted APs positions are estimated by hierarchal inferencing and the candidate APs positions are the positions estimated maxima detection. In the fused RSSI field map, we represent the final estimated positions based on convex-hull alignment. Note: the fused RSSI fields are represented for detailed illustration, our method predicts RSSI fields for each AP separately. The inferior convex hulls are the convex hulls with poor alignment error.}}
 \label{Exp_fields}
\end{figure*}

\begin{figure}[t]
\centering
 \includegraphics[width=0.99\linewidth]{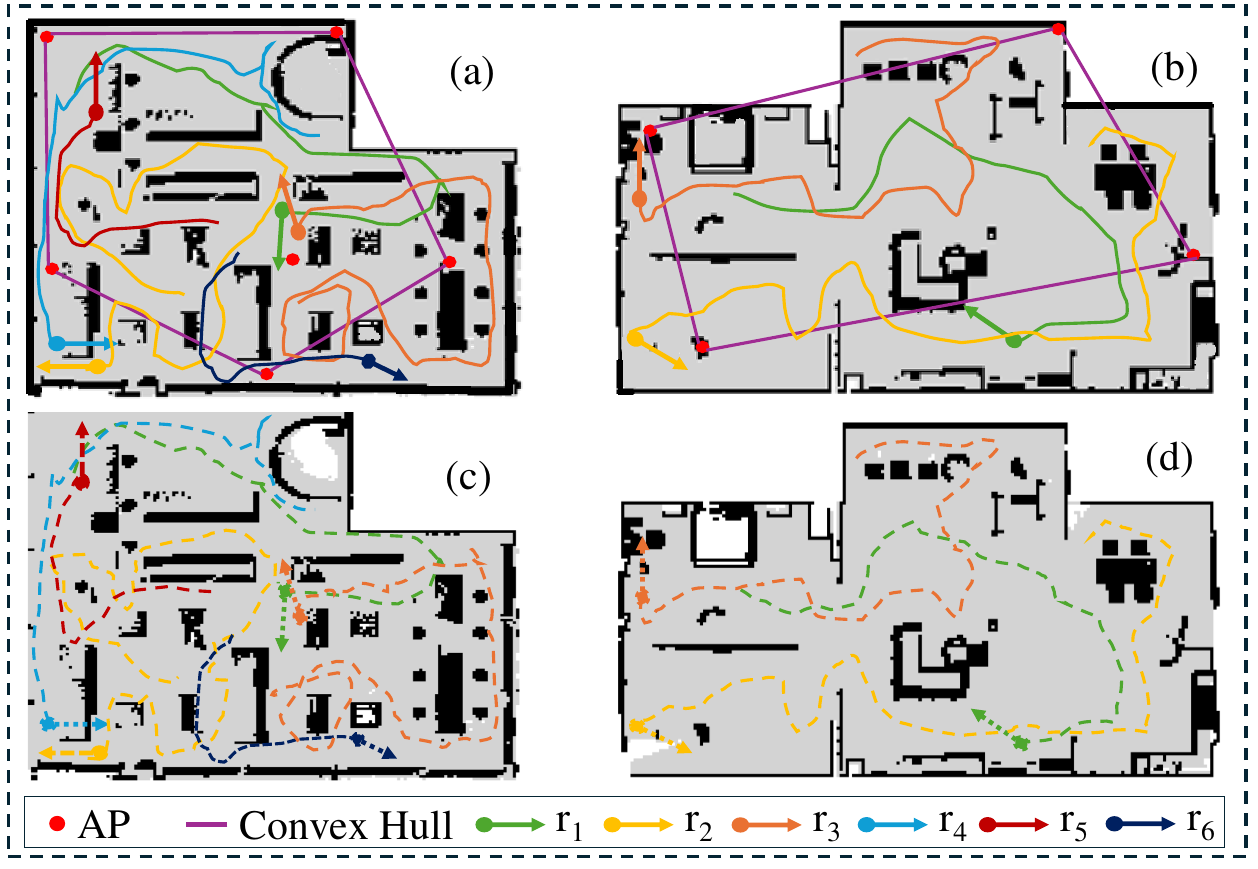}
 \caption{\footnotesize{Sample trial exploration outcomes in AWS Bookstore (a, c) and House (b, d) worlds. (a), (b) represents ground truth maps with true trajectories of robots and formed convex hull of APs. (c), (d) represents the MGPRL relative localization trajectories (translated into global frame). }}
 \label{Exp_explore}
\end{figure}

\begin{figure*}[t]
\centering
 \includegraphics[width=0.99\linewidth]{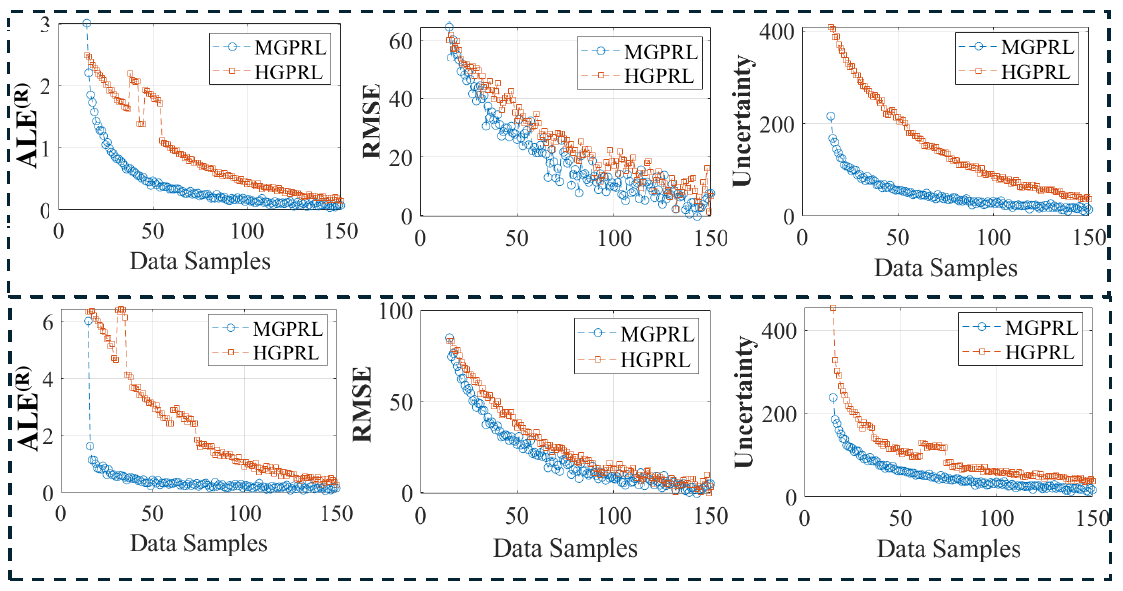}
 \caption{\footnotesize{The temporal analysis of ALE of robots' relative localization, RMSE and uncertainty
(averaged for all AP fields) of co-regionalized MOGP with respect to number of samples collected in AWS house (\textbf{Top}) and bookstore (\textbf{Bottom}).}}
 \label{temporalanalysis}
\end{figure*}

\subsection{Co-regionalized MOGP \& AP localization}
We evaluate the performance of the co-regionalized MOGP by measuring\footnote{These performance metrics are calculated based on average performance of all RSSI fields ($n \times m$) from all APs ($m$) across all robots ($n$). The $\text{ALE}^{(AP)}$ and $\text{ALE}^{(R)}$ represent the average ALE of all APs and robots, respectively.} prediction RMSE, uncertainty, training time, and inference time. Moreover, we assess the AP localization performance using absolute localization error ($\text{ALE}^{(AP)}$) under various noise levels as shown in Fig. \ref{Exp_Box}. Since we utilize existing hierarchical inferencing ($V=4$) for initial AP localization, we utilize the re-estimated AP candidate positions after the convex-hull alignment. The ALE of AP localization of MGPRL is improved by 54.38\% and 49.7\% in house and bookstore worlds, respectively. In terms of average increase in $\text{ALE}^{(AP)}$, with increasing noise measurement collection (noise levels), MGPRL is 2.97 and 1.45 times the HGPRL, making the proposed method more resistant to noise. Additionally, the RMSE and uncertainty of RSSI field prediction with MOGP reduced by 33.45\% and 57.33\% in comparison with HGPRL. This further demonstrates that MOGP provides more robust uncertainty estimates than GP, leading to improved reliability of our relative localization. Moreover, the training time and inference time of co-regionalized MOGP is approximately 34.44\% and 37.04\% lower than GP in HGPRL. In our 15 trails of experiments, MGPRL exhibited approximately 43\% lesser deviation in each trail over HGPRL.

\subsection{Relative Localization}
To analyze the performance of relative localization, we measure the ALE of the robots ($\text{ALE}^{(R)}$) with respect to the ground truth, extracted from ROS Gazebo. The predicted and fused RSSI fields for all three robots and four APs in the House experiment are shown in Figure \ref{Exp_fields}. 
Figure \ref{Exp_explore} presents 2D maps of sample trial explorations, illustrating convex hull formations and trajectory comparisons of robots in the AWS house and bookstore environments. Not all APs contribute to the convex hull. In the bookstore experiment (a), one AP was not included in convex hull formation, so the robot system relied on the remaining five APs for hull alignment, eliminating the need for additional RSSI sampling of AP inside the hull.  Hierarchical inferencing achieved an accuracy of 73.6\%, which led to inaccuracies in relative localization using HGPRL. For example, in the $AP_4$ field for $r_1$ (Fig. \ref{Exp_fields}), the AP was incorrectly estimated by hierarchical inferencing; this error was subsequently detected as a local maximum and corrected by the convex-hull alignment method in MGPRL. Additionally, in the RSSI fields of $AP_1$ and $AP_3$ for $r_3$, significant corrections are performed by MGPRL in AP localization. Overall, MGPRL successfully corrected 93.5\% of the incorrect AP estimates (which occurred in 26.4\% of RSSI fields), thereby ensuring robust relative localization.

The performance results of relative localization through temporal analysis of ALE of robots, RMSE, and uncertainty of MOGP are shown in Fig. \ref{temporalanalysis}. The state-of-the-art HGPRL showed abrupt spikes in the ALE of robot positions (see Fig. \ref{temporalanalysis}) during both the house and bookstore experiments. These spikes, lasting for a short duration, presented an inaccurate pose estimation (by hierarchical inference) of AP due to the presence of multiple maxima and uncertainty. The proposed MGPRL is able to perform accurate and robust relative localization throughout the exploration. Moreover, MGPRL converges to the least $ALE^{(R)}$ and uncertainty with less number of samples when compared to HGPRL. As the number of robots increased from $n=3$ to $6$, the MGPRL $ALE^{(R)}$ is increased by 22.3\%, whereas HGPRL's  $ALE^{(R)}$ is increased by 78.3\%. Overall, MGPRL achieved 73\% and 84\% improvements in relative localization ($ALE^{(R)}$) over HGPRL in house and bookstore worlds, respectively.

\begin{figure*}[t]
\centering
 \includegraphics[width=0.99\linewidth]{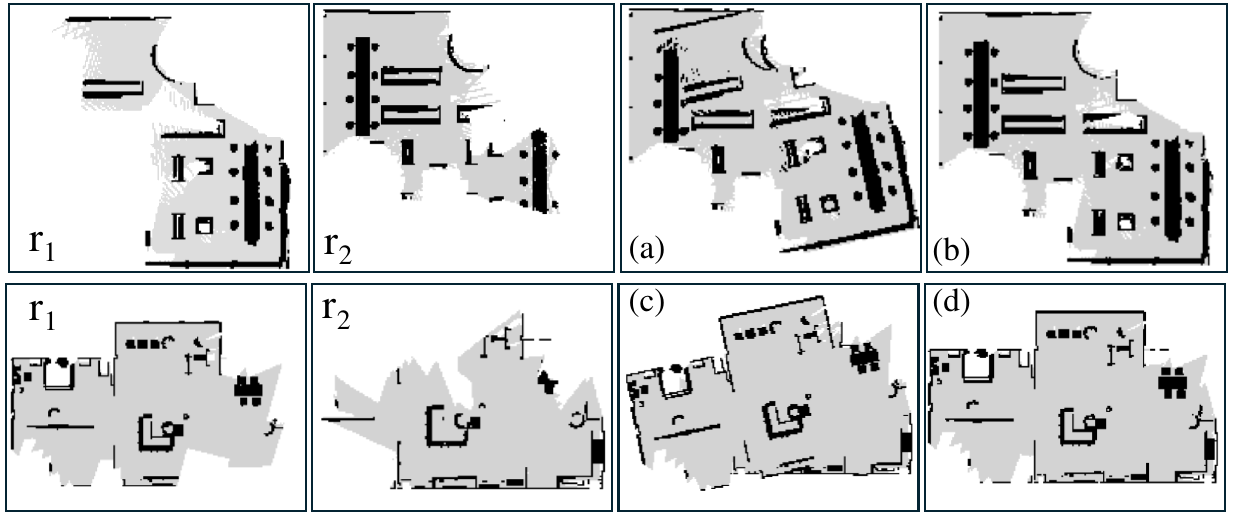}
 \caption{\footnotesize{Comparison of multi-robot grid map merging using a feature-based approach \cite{horner2016map} (a,c) and MGPRL (b,d) in the AWS house (\textbf{Bottom}) and bookstore  (\textbf{Top}). MGPRL helped bootstrap the merging process for increased accuracy and computation performance.}}
 \label{gridmapmergin}
\end{figure*}

\subsection{MGPRL-based Grid Map Merging}
To analyze the effectiveness of MGPRL, we performed relative localization–based map merging of exploration maps, as shown in Fig. \ref{gridmapmergin}. To achieve this, we combined the maps (collected in the middle of exploration), where there are only few regions of overlap between the maps. This process is compared with widely-used existing state-of-the-art ORB feature-based map merging approach \cite{horner2016map}. For effective comparison, we utilize structural similarity index measure (SSIM) with the ground-truth map. The proposed MGPRL merging is 35\% and 28\% effective over feature based merger in AWS house and bookstore respectively. Moreover, MGPRL (requiring only relative localization) reduced the computational time by approximately a factor of eight compared to the feature-based merger, taking  $0.79 \pm 0.18$ seconds instead of $6.3 \pm 1.27$ seconds. Overall, these results demonstrate that MGPRL not only provides accurate and robust relative localization, but also enhances the efficiency and quality of map merging.

\begin{figure*}[t]
\centering
 \includegraphics[width=0.99\linewidth]{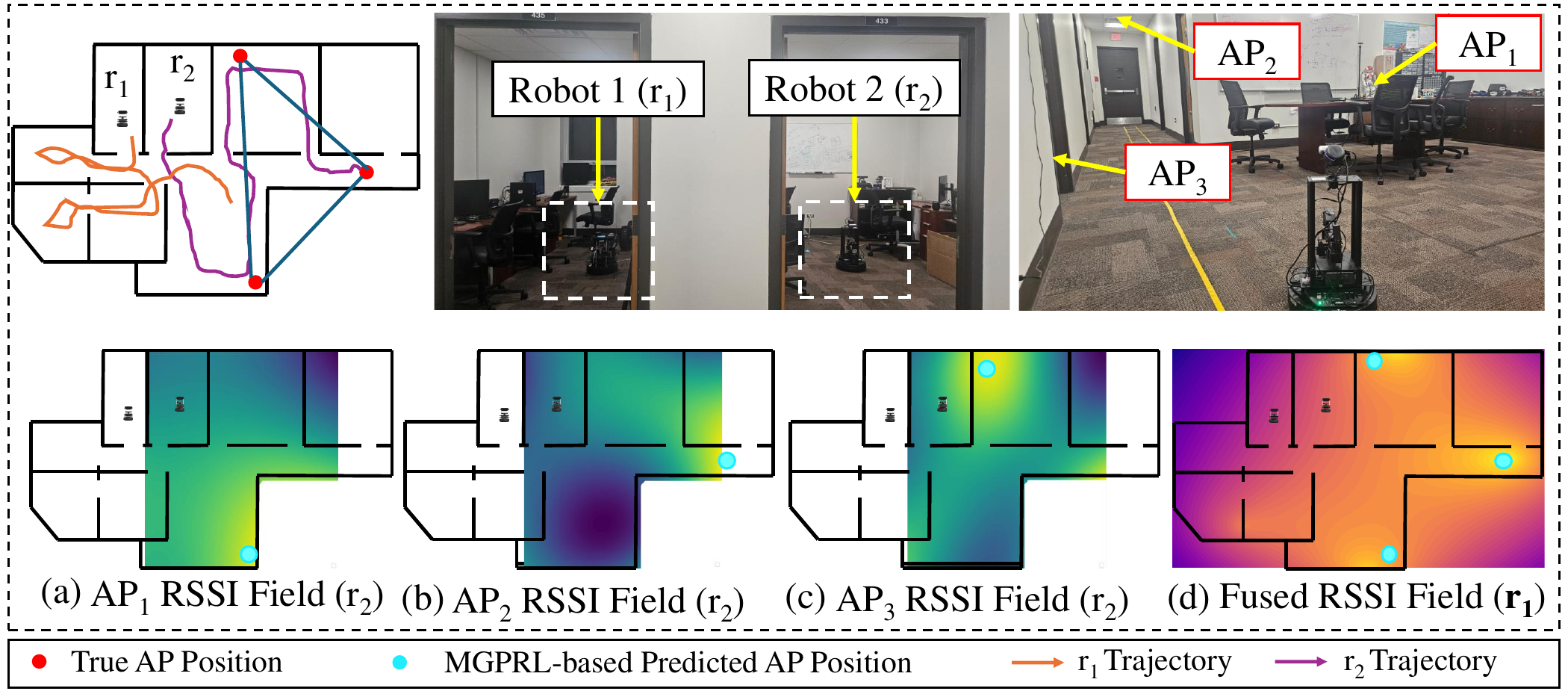}
 \caption{\footnotesize{A trail of multi-robot exploration experiment conducted in real-word utilizing MGPRL. The top row represents the ground truth map and the initial states of the robots. The bottom row depicts the collected RSSI fields of $r_2$ (a,b,c) for each AP and the fused RSSI field of $r_1$ (d). The RSSI fields are illustrated within the environment for demonstration. }}
 \label{real}
\end{figure*}

\subsection{Real-World Experiments}
To assess practicality and adaptability, we deployed the MGPRL distributed framework with ROS-Noetic to execute a multi-robot exploration using two TurtleBot2e across a $10m \times 13m$ multi-room laboratory setting. Both the robots are configured and connected to three Wi-Fi sources (APs). The robots were initially positioned in separate rooms without visibility to each other or the APs, operating under non-line-of-sight conditions in the RSSI map. We conducted four trials, during which both robots successfully completed their exploration. A sample trail of the exploration is shown in Fig. \ref{real}. The robots were able to determine the location of the APs with an accuracy of $0.32\pm 0.4$m (outperforming HGPRL ($0.5 \pm 0.12$m)), and were able to perform relative localization efficiently. The experiment confirmed the effectiveness of the proposed MGPRL in tackling real-world challenges involving noisy RSSI measurements, non-line-of-sight environments. Additionally, it highlights MGPRL’s adaptability to various multi-robot tasks, including rendezvous and coordinated formation. The MGPRL is particularly suitable for scenarios where direct visibility between robots is absent or where their paths do not intersect—situations (disjoint environments)  where feature-based localization relying on loop closure would be impractical.

\section{Conclusion}
This paper presents MGPRL, a distributed framework for multi-robot relative localization using coregionalized MOGP for efficient RSSI-based AP localization. By incorporating weighted convex-hull alignment, MGPRL improves accuracy in noisy, multi-peak environments, outperforming state-of-the-art methods with up to 52\% and 73\%  improvement in AP localization and relative localization respectively. This work can be extended by developing a framework that estimates AP positions using a single RSSI field, reducing computational costs. Additionally, it can be further improved for relative localization with anonymous AP predictions based on the multi-AP RSSI distributions.

\section*{Acknowledgements}
Research was sponsored by the U.S. Army Research Laboratory and was accomplished under Cooperative Agreement Number W911NF-17-2-0181 (DCIST-CRA). 
The views and conclusions contained in this document are those of the authors and should not be interpreted as representing the official policies, either expressed or implied, of the Army Research Laboratory or the U.S. Government. The U.S. Government is authorized to reproduce and distribute reprints for Government purposes notwithstanding any copyright notation herein.

\bibliographystyle{IEEEtran}
\bibliography{Main}

\end{document}